%% file: main.tex
\pdfoutput=1

\documentclass[11pt]{article}

\usepackage[]{acl}

\usepackage{times}
\usepackage{latexsym}

\usepackage{booktabs}
\usepackage{siunitx}
\usepackage{amsmath,amssymb,amsthm}
\usepackage{multirow}

\usepackage[T1]{fontenc}

\usepackage[utf8]{inputenc}

\usepackage{microtype}
\usepackage{subfig}
\usepackage{tabularx}

\newcommand{\colorcorrect}{\setlength{\fboxsep}{1pt}\colorbox{green!40}}

\usepackage{graphicx}

\title{Learning to Segment Words for Low-Resource Language Modelling}
\title{Subword Segmental Language Modelling for Nguni Languages}

\author{Francois Meyer and Jan Buys\\
  Department of Computer Science \\
  University of Cape Town \\
  \texttt{MYRFRA008@myuct.ac.za, jbuys@cs.uct.ac.za}}

\begin{document}
\maketitle
\begin{abstract}
Subwords have become the standard units of text in NLP, enabling efficient open-vocabulary models.
With algorithms like byte-pair encoding (BPE), subword segmentation is viewed as a preprocessing step applied to the corpus before training. This can lead to sub-optimal segmentations for low-resource languages with complex morphologies.
We propose a subword segmental language model (SSLM) that learns how to segment words while being trained for autoregressive language modelling. 
By unifying subword segmentation and language modelling, our model learns subwords that optimise LM performance.
We train our model on the 4 Nguni languages of South Africa. These are low-resource agglutinative languages, so subword information is critical. 
As an LM, SSLM outperforms existing approaches such as BPE-based models on average across the 4 languages. Furthermore, it outperforms standard subword segmenters on unsupervised morphological segmentation.
We also train our model as a word-level sequence model, resulting in an unsupervised morphological segmenter that outperforms existing methods by a large margin for all 4 languages.
Our results show that learning subword segmentation is an effective alternative to existing subword segmenters, enabling the model to discover morpheme-like subwords that improve its LM capabilities.
\end{abstract}

\input{1intro}

\input{2subword_segmentation}

\input{3sslm}

\input{4lm}

\input{5ums}

\input{6analysis}

\input{7related_work}

\input{8conclusion}

\input{9limitations}

\bibliography{anthology,custom}
\bibliographystyle{acl_natbib}

\clearpage

\input{appendix}

\end{document}

%% file: 1intro.tex
\section{Introduction}


Subword segmentation has become a standard practice in Natural Language Processing (NLP). The dominant approach is to run an algorithm like BPE \citep{sennrich-etal-2016-neural} as a preprocessing step, segmenting the corpus into subwords. This enables the model to learn features based on subwords, compose words, and handle rare and unknown words as an open-vocabulary model. 
Subword segmentation is an active area of research, since no single technique outperforms others across all tasks, languages, and dataset sizes \cite{zhu-etal-2019-importance,zhu-etal-2019-systematic}.
Besides deterministic segmenters like BPE, stochastic algorithms like unigram LM (ULM) \citep{kudo-2018-subword} have also been proposed. 


\begin{figure}[t]
    \centering
	\begin{tabular}{lc} 
		\toprule
	     \multicolumn{2}{c}{\textbf{sesihambe}}\\
	     \midrule
	     Morphemes & se-si-hamb-e  \\
	     BPE &  sesi-ha-mbe\\
	     Unigram LM & se-si-hambe\\
	     Morfessor & se-s-ihambe\\
	     SSLM & se-si-hamb-e\\
		\bottomrule
	\end{tabular}
	\captionof{table}{Segmentations of the isiXhosa word \emph{sesihambe} produced by existing subword segmentation algorithms, compared to the actual morphemes and the output of our model (SSLM).} 	\label{examples}
\end{figure}

Subword segmentation is particularly important for the Nguni languages of South Africa (isiXhosa, isiZulu, isiNdebele, and Siswati) because they are agglutinative languages that are  written conjunctively.\footnote{The Sotho-Tswana languages of South Africa are also agglutinative, but are written disjunctively i.e. a single linguistic word may be written as several orthographic words.}
These are morphologically rich languages in which words are formed by stringing together morphemes  \cite{taljard-bosch-2006-comparison}. 
Morphemes are the primary linguistic units. 
For example, the isiXhosa word ``sesihambe'' means ``we are gone'', where ``se'' means ``we'',
``si'' means ``are'', and ``hamb-e'' means ``gone'', with the ``-e'' suffix indicating past tense.
As shown in Table \ref{examples}, existing segmenters do not reliably capture this.

The Nguni languages are under-resourced, which compounds the importance of subword segmentation.
Available datasets are small, so any held-out dataset will contain rare or previously unseen words. Therefore it is critical for models to learn useful subword features and effectively model morphological composition. 
In a low-resource setting it may then be more effective to learn subword segmentation as part of model training rather than as a distinct preprocessing step.


The probabilistic models underlying existing subword segmentation methods such as ULM and Morfessor \cite{creutz-etal-2007-morfessor} assume that subwords are context-independent, making them unsuitable for language modelling. 
In this paper we propose a subword segmental language model that simultaneously learns how to segment words  while training as an autoregressive LM. This allows the model to learn subword segmentations that optimise a left-to-right language modelling objective, thereby being conditioned on the context. Our model learns the subwords that it can most effectively leverage for language modelling.



We train our model in the 4 Nguni languages of South Africa. We compile LM datasets for these languages from publicly available corpora and release our train/validation/test sets.
On intrinsic language modelling performance averaged across the 4 languages our model outperforms neural LMs trained with characters, BPE, and ULM subwords.
On the task of unsupervised morphological segmentation (which determines to what extent subwords correspond to actual morphemes) our model outperforms standard subword segmenters like BPE and ULM on average across the 4 languages. 

In addition to these LMs, we train a second set of subword segmental models that train on single words in isolation (without having to model context for long-range language modelling).  Our word-level models outperform all existing methods on unsupervised morphological segmentation (including segmenters like Morfessor) by a large margin across all 4 languages.
Finally, we discuss the importance of a subword lexicon to our model, analysing how hyperparameters that control lexicon construction affect performance. 
In summary, this paper makes the following contributions:\footnote{Our code, trained models, and datasets are available at \url{https://github.com/francois-meyer/subword-segmental-lm}.}
\begin{enumerate}
    \item We propose a subword segmental language model (SSLM) that unifies subword segmentation and language modelling in a single end-to-end neural architecture.
    \item We compile and release curated LM datasets for 4 Nguni languages.
    \item We evaluate our model as an LM and an unsupervised morphological segmenter, and it outperforms existing methods on both tasks. 
    \item We present an analysis of how lexicon-related hyperparameters affect our model.
\end{enumerate}



%% file: 2subword_segmentation.tex
\begin{figure*} 
  \includegraphics[width=\textwidth]{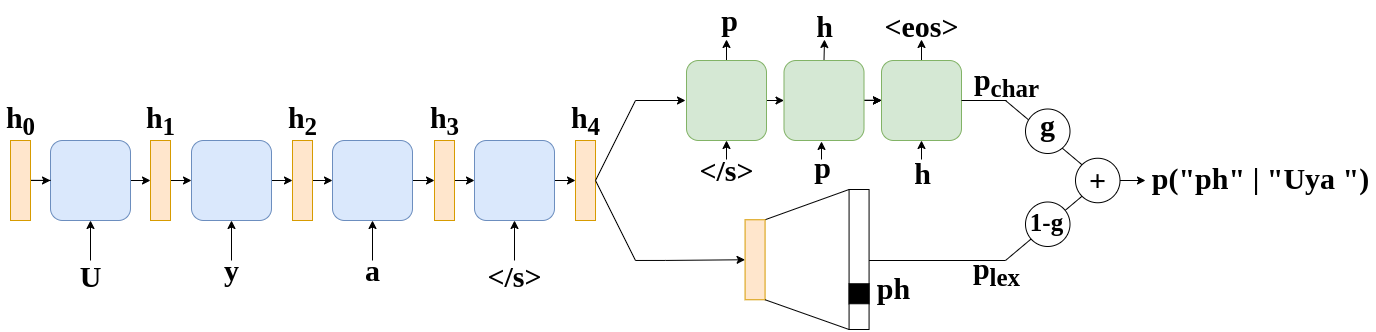}
  \caption{The SSLM computing the probability for the subword segment ``ph'' in the isiXhosa sentence ``Uya phi?''  
  A character-level LSTM encodes the unsegmented text history ``Uya '', while a mixture model (equation \ref{mixture}) that interpolates between a separate character-level LSTM decoder and a lexicon model generates the segment ``ph''. This is repeated for all possible subwords in a sequence to compute the forward scores  of equation \ref{marginal_markov}.}
  \label{encoder}
\end{figure*}

\section{Subword Segmentation}

In this section we review the paradigm that currently dominates subword segmentation, discuss its limitations, and introduce the family of models we draw inspiration from for our approach to subword segmentation --- segmental sequence models.


\subsection{Subword Segmentation Algorithms}

Recently proposed subword segmentation algorithms 
start with some initial vocabulary (e.g. all characters) and iteratively amend it based on corpus subword statistics until a pre-specified vocabulary size has been reached. 
The goal of BPE \cite{sennrich-etal-2016-neural} is to represent common characters sequences as distinct vocabulary items.
ULM \cite{kudo-2018-subword} aims to maximise the likelihood of the training corpus under a unigram LM, in which subwords are generated independently.  

These algorithms work well in certain contexts, but are not universally applicable.
\citet{klein-tsarfaty-2020-getting} show that they are sub-optimal for morphologically rich languages.
\citet{zhu-etal-2019-systematic} show that the best method varies across languages and tasks, and  existing segmenters require extensive tuning. 
They also find that subword segmentation is particularly beneficial for low-resource languages,
but on average a simple character n-gram method outperforms BPE \cite{zhu-etal-2019-importance}.  

Recently it has become popular to construct shared multilingual vocabularies, but this leads to over-segmented words in low-resource languages  \citep{wang-etal-2021-multi-view, acs-2021-exploring}.
Some have argued that these problems can be overcome by avoiding segmentation altogether \citep{DBLP:journals/corr/abs-2103-06874} or by more sophisticated hyperparameter tuning \citep{salesky-etal-2020-optimizing}.
But the limitations arise partly from the fact that the segmentation algorithms themselves are separated from model training.
To overcome this we turn to a different paradigm, where we can cast subword segmentation as something for the model to learn.

\subsection{Segmental Sequence Models}

The main idea behind segmental sequence modelling is to let the model learn segmentation itself. This involves treating sequence segmentation as a latent variable to be marginalised over. The motivation behind this is that the model would be able to ``discover'' the optimal segments for sequence prediction. These segments might correspond to natural underlying sequence units, such as words in text or phonemes in speech.

Variants of this idea have been used in a few neural sequence models.
\citet{kong-etal-2015-segmental} propose a bidirectional RNN that learns segment embeddings for handwriting recognition and POS tagging. \citet{wang-etal-2017-sequence} propose SWAN (Sleep-WAke Networks), a segmental RNN for text segmentation and speech recognition. Both of these models use dynamic programming to efficiently compute marginal likelihood during training (by summing over all possible segmentations) and to find the most likely segmentation of a sequence.

\citet{sun-deng-2018-unsupervised} coined the term ``segmental language model'' (SLM) in applying this approach to Chinese language modelling for unsupervised  word segmentation. 
\citet{kawakami-etal-2019-learning} extended their approach by equipping the model with a lexical memory and introducing segment length regularisation. 
Segmental models for word discovery have also been proposed as masked LMs \citep{downey-etal-2021-masked} and bi-directional LMs \citep{wang-etal-2021-unsupervised}.
Inspired by these works, we adapt segmental sequence modelling for the joint task of language modelling and subword discovery.


%% file: 3sslm.tex
\section{Subword Segmental Language Model}

Our SSLM combines autoregressive language modelling and subword segmentation in a single model that can be trained end-to-end. The architecture is shown in Figure \ref{encoder}. 
It represents a radical divergence from segmenters like BPE and ULM, which view subword segmentation as context-independent. The SSLM views subword segmentation and language modelling jointly, so it can learn subwords that optimise conditional LM generation.

\subsection{Generative Model}


The SSLM generates a sequence of space-separated words $\mathbf{w} = \mathbf{w_1}, \mathbf{w_2}, \dots, \mathbf{w_n}$, corresponding to an underlying sequence of characters $\mathbf{x}$, and generates each word $\mathbf{w_i}$ as a sequence of subwords $\mathbf{s_i} = s_{i1}, s_{i2}, \dots, s_{i|\mathbf{s_i}|}$. The probability of a text sequence $\mathbf{w}$ is computed through the marginal distribution over all possible word segmentations as
\begin{align} \label{marginal} 
    p(\mathbf{w}) = \sum_{\mathbf{s}: \pi(\mathbf{s}) = \mathbf{w}} p(\mathbf{s}),
\end{align}
where $\pi(\mathbf{s})$ is the unsegmented text underlying the sequence of segmented words $\textbf{s} = \mathbf{s_1}, \mathbf{s_2}, \dots, \mathbf{s_n}$. Using the chain rule, we define the probability of a sequence of segmented words as
\begin{align} 
    p(\mathbf{s}) = \prod_{i=1}^{|\mathbf{w}|} \prod_{j=1}^{|\mathbf{s_i}|} p(s_{ij} | \mathbf{s_{\leq i, <j})}, 
\end{align}
where $\mathbf{s_{\leq i, <j}}$ is the subword sequence preceding the $j^{\mathrm{th}}$ subword of the $i^{\mathrm{th}}$ word (this includes all subwords in the preceding words and the subwords preceding $s_{ij}$ in the current word). 

We treat white spaces and punctuation as assumed segments that are equivalent to 1-character words. In this way we implicitly model the end of a word. When the model predicts a non-alphabetical character (e.g. space) that is equivalent to a word boundary. Segments cannot cross word boundaries, so the only segmentation learned by the model is how to segment words into subwords.

\begin{figure*}[t]
    \centering
	\begin{tabular}{lcccccccccc} 
		\toprule
		 & \multicolumn{5}{c}{\textbf{Validation set BPC}} &
		 \multicolumn{5}{c}{\textbf{Test set BPC}} \\
		 \cmidrule(lr){2-6} \cmidrule(lr){7-11}
		 \textbf{Model} & \textbf{xh} & \textbf{zu} & \textbf{nr} & \textbf{ss} & \textit{avg} & \textbf{xh} & \textbf{zu} & \textbf{nr} & \textbf{ss} & \textit{avg}\\
	     \midrule
	     Char-LSTM& 1.24  & 1.22& 1.41 & \textbf{1.38} & 1.31& 1.32&  1.26& 1.39& 1.30&1.32\\
	     BPE-LSTM & 1.23  & \textbf{1.19}&  1.39 & \textbf{1.38} & 1.30& 1.30 & 1.22& 1.39& \textbf{1.28} & 1.30\\
	     ULM-LSTM & \textbf{1.22}& 1.23 & 1.39 & 1.40 & 1.31& \textbf{1.25}& 1.27 & 1.39& 1.31& 1.31\\
	     \midrule
	     Char-Transformer & 1.51& 1.43&1.49 & 1.49 & 1.48& 1.53 & 1.48 & 1.47& 1.43&1.48\\
	     BPE-Transformer & 1.30& 1.22& 1.38& \textbf{1.38}&1.32 & 1.33& 1.27& 1.36& 1.30&1.31\\
	     ULM-Transformer & 1.32& 1.22& 1.38& \textbf{1.38}& 1.35& 1.34 & 1.27& 1.36& 1.29& 1.31\\
	     \midrule
	     SSLM & 1.24 & \textbf{1.19} & \textbf{1.35} & \textbf{1.38} & \textbf{1.29}& 1.27& \textbf{1.21}& \textbf{1.35} & \textbf{1.28} &\textbf{1.28}\\
		\bottomrule
	\end{tabular}
	\captionof{table}{Intrinsic LM performance, as measured by BPC scores on the validation and test sets.} 	\label{lm_results}
\end{figure*}

\subsection{Dynamic Programming Algorithm}

Conditioning the probabilities of a segment $p(s_{ij} | \mathbf{s_{\leq i, <j}})$ on all possible segmentation histories is computationally intractable, so we follow previous segmental sequence models by introducing a conditional semi-Markov assumption,
\begin{align} \label{marginal_markov}
    p(s_{ij} | \mathbf{s_{\leq i, <j})} &\approx p(s_{ij} | \pi(\mathbf{s_{\leq i, <j}})) \\
    & = p(s_{ij} | \mathbf{x_{<k}}),
\end{align}
where $\mathbf{x_{<k}}$ is the raw, unsegmented text preceding the current segment (assuming the current segment starts at the $k^{\mathrm{th}}$ character). Now the segment generation probability does not depend on the segmentations in the preceding sequence of words, or within the current word. Instead, the probability is conditioned on the unsegmented word and character history. This enables us to compute the marginal likelihood of equation \ref{marginal} incrementally using a dynamic programming algorithm. Given $\alpha_0 = 1$, at each step the algorithm computes a forward score,
\begin{align} \label{dp}
    \alpha_t = \sum_{k= f(\mathbf{x}, t)}^{t} \alpha_k p(s = \mathbf{x_{k:t}} | \mathbf{x_{<k}}), 
\end{align}
where $k$ is the starting index of the current word (the longest possible subword segment is the entire word). Each of the expressions in the summation represents the probability of concluding the sequence at character $t$ by generating a segment starting at character $k$. We can efficiently compute the marginal in equation \ref{marginal} as $p(\mathbf{w}) = p(\mathbf{x}) = \alpha_{|\mathbf{x}|}$.


\subsection{Neural Model}

Each segment probability is computed as a mixture of 2 probability distributions computed as
\begin{align} \label{mixture}
    p(s_{ij} | \mathbf{x_{<k}}) =\,\, & g_k p_{\mathrm{char}} (s_{ij} | \mathbf{h_{k}}) + \nonumber\\
    & (1-g_k) p_{\mathrm{lex}} (s_{ij} | \mathbf{h_{k}}),
\end{align}
where $\mathbf{h_k}$ encodes the sequence history, $g_k$ is a mixture coefficient, $p_{\mathrm{lex}}$ is a fully connected neural layer that generates the entire segment from a lexicon as a single event, while $p_{\mathrm{char}}$ is an LSTM that generates the segment character by character. The probability assigned by $p_{\mathrm{char}}$ is computed using the chain rule over the character sequence followed by a special end-of-segment \texttt{<eos>} character.

\begin{figure}[t] 
    \centering
	\begin{tabular}{lcc} 
		\toprule
	     & \multicolumn{2}{c}{\textbf{\# Tokens}}\\
	     \cmidrule(lr){2-3}
	     \textbf{Language} & \textbf{Train} & \textbf{Valid/Test}\\
	     \midrule
	     isiXhosa (xh)& 3.4mil & 190k \\
	     isiZulu (zu) & 3.1mil & 200k \\
	     isiNdebele (nr) & 450k & 25k\\
	     Siswati (ss) & 500k & 28k \\
		\bottomrule
	\end{tabular}
	\captionof{table}{Language modelling dataset sizes.} 	\label{dataset_sizes}
\end{figure}

The character and lexicon models are conditioned on the unsegmented text history through $\mathbf{h_k}$, a vector representation computed by a character-level LSTM encoder. The LSTM \citep{hochreiter-schmidhuber-1997-long} is better suited to the probabilistic conditioning required for our model than the Transformer \citep{vaswani-etal-2017-attention}, since it computes a single hidden state $\mathbf{h_k}$ representing the entire sequence history. This  can be passed to segment predictors within an efficient dynamic programming algorithm.
The mixture coefficient $g_k$ is also computed from $\mathbf{h_k}$ with a fully connected neural layer, so the model can learn when to rely on the lexicon and when to revert to character-by-character generation.
The model is trained by maximising the log likelihood over the training corpus.

\begin{figure*}[t]\small
    \centering
	\begin{tabular}{lcccccccccccc} 
		\toprule
	     & \multicolumn{3}{c}{\textbf{xh}} & \multicolumn{3}{c}{\textbf{zu}} & \multicolumn{3}{c}{\textbf{nr}} & \multicolumn{3}{c}{\textbf{ss}} \\
	     \cmidrule(lr){2-4} \cmidrule(lr){5-7} \cmidrule(lr){8-10} \cmidrule(lr){11-13}
	     \textbf{Model} & P & R & F1 & P & R & F1 & P & R & F1 & P & R & F1\\
	     \midrule
	     \textbf{BPE} 
	     & 25.00 & 25.42 & 25.21 &
         25.20 & 22.89 & 23.99  &
         21.23 & 21.70 & 21.46 &
         22.30 & 24.38 & 23.30

\\ 
         \textbf{ULM} 
	     & 33.01 & 34.07 & 33.53
         & 29.04 & 27.07 & 28.02
         & 25.88 & 26.56 & 26.21
         & 25.70 & 29.25 & 27.36

\\
	     
	     \textbf{Morfessor} & 
	     21.68 & 17.05 & 19.09
        & 20.25 &17.58 & 18.82
        & 18.52 & 17.70 & 18.10
        & 24.02 & 22.14 & 23.04

\\    
	     \midrule
	     \multicolumn{13}{l}{\textbf{Entropy-based (Stddev)}} \\
	     \midrule
	     LSTM & 
	     40.99 & 30.43 & 34.93
 & 39.26 & 28.23 & 32.85
 & 38.61 & 29.01 & 33.13
 & 33.95&30.40&32.07
\\

	     Transformer & 
	     42.99 & 33.92 & 37.92&  39.24 & 28.93 &33.31&
	     38.93 &29.65 &33.66 &
	     33.16&29.74&31.35\\
	     \midrule
	     
	     \multicolumn{13}{l}{\textbf{Subword segmental models}} \\
	     \midrule
	     SSLM & 
	     20.43 & \textbf{41.86} & 27.46 & 
	     24.03 & \textbf{43.69} & 31.01&
	     26.44 &\textbf{44.59} &	33.20& 
	     19.56 & 30.12 & 23.71  \\
	     Word-level 
	     & \textbf{44.55} & 38.07 & \textbf{41.06}
	     & \textbf{49.44} & 39.75 & \textbf{44.07}
	     & \textbf{41.39} & 38.01 & \textbf{39.63}
	     & \textbf{38.34} & \textbf{38.26} & \textbf{38.30}   \\
		\bottomrule
	\end{tabular}
	\captionof{table}{Morpheme identification  (MI) metrics averaged over the annotated evaluation set.} 	\label{ums1}
\end{figure*}	

\section{Language Modelling} \label{sect4_lm}

We evaluate our SSLM on intrinsic language modelling performance to determine whether learning subword segmentation during training can improve the inherent predictive capabilities of an LM.


The lexicon is constructed before training. It contains all subwords shorter than a prespecified maximum length $L$ that occur in the training corpus. The lexicon size $V$ is also prespecified. The lexicon consists of the $V$ most frequent subwords up to $L$ characters long.
The lexicon model $p_{lex}$ outputs a probability for each subword in the lexicon.


%% file: 4lm.tex
\subsection{Data} \label{datasubsction}

We train our models on LM datasets we compiled for isiXhosa (\textbf{xh}), isiZulu (\textbf{zu}), isiNdebele (\textbf{nr}), and Siswati (\textbf{ss}). 
For each language we collected publicly available datasets and combined them into a single corpus. To avoid some of the pitfalls of low-resource data collection \cite{kreutzer-etal-2021-quality}, we set specific criteria for including datasets. We collected datasets from reputable sources such as the South African Centre for Digital Language Resources (SADiLaR).\footnote{\url{https://repo.sadilar.org/discover}}
We also inspected individual datasets (manually and using scripts) and discarded datasets of questionable quality (e.g. containing a significant amount of English text). 
The sources we used are listed in the appendix. 
We split our corpora 80\%/10\%/10\% into train/validation/test sets. The dataset sizes are listed in Table \ref{dataset_sizes}. 

\subsection{Evaluation}

We evaluate our models using bits-per-character (BPC) - an intrinsic evaluation metric that measures how well an LM predicts a corpus. It is cross-entropy-based and normalised by character length, so it allows for comparison across different subword segmentations. BPC is computed as
\begin{equation}
\mathrm{BPC}(X) = - \frac{1}{N} \sum_{\mathbf{x} \in X} \log_2 p(\mathbf{x}), 
\end{equation}
where $X$ is a corpus of sequences $\mathbf{x}$ and $N$ is the length of the corpus in characters. 

\subsection{Models and Training}

For each language we train an SSLM and 6 baselines. Our baselines use 3 standard subword methods: character tokens, BPE, and ULM. For each we train LSTM and Transformer LMs.
We tune the hyperparameters of all our models by optimising for validation set BPC. The hyperparameter settings for our SSLMs and baselines are provided in appendix \ref{appendix_hyperparameters}.

\subsection{Results}

BPC scores on the validation and test sets are shown in Table \ref{lm_results}.
The SSLM emerges as the best LM on average across the languages. 
It achieves the best BPC scores for all languages except isiXhosa, where it still comes very close to the best-performing model.
Among the baselines the best neural architecture and subword segmenter depends on the language. 
For example, the Transformer subword models outperform the LSTM models for isiNdebele, but perform surprisingly poorly for isiXhosa.
This inconsistency is one of the primary limitations we are trying to address with our approach. The results show that the SSLM succeeds in this regard - it is more consistent and generally applicable across these languages. 

%% file: 5ums.tex
\begin{figure*}[t]\small
    \centering
	\begin{tabular}{lcccccccccccc} 
		\toprule
	    & \multicolumn{3}{c}{\textbf{xh}} & \multicolumn{3}{c}{\textbf{zu}} & \multicolumn{3}{c}{\textbf{nr}} & \multicolumn{3}{c}{\textbf{ss}} \\
 	     \cmidrule(lr){2-4} \cmidrule(lr){5-7} \cmidrule(lr){8-10} \cmidrule(lr){11-13}
	     \textbf{Model} & P & R & F1 & P & R & F1 & P & R & F1 & P & R & F1\\
	     \midrule
	     \textbf{BPE} 
	     & 40.71 & 41.79 & 41.24
         & 45.23 & 39.00 & 41.88
         & 38.43 & 39.76 & 39.08
         & 32.73 & 37.75 & 35.06

\\ 
         \textbf{ULM} 
	     & 45.92 & 48.24 & 47.05
         & 47.04 & 42.25 & 44.52
         & 41.48 & 43.18 & 42.31
         & 35.91 & 44.04 & 39.56

\\
	     
	     \textbf{Morfessor} 
	     & 38.14 & 25.46 & 30.54
	     & 38.83 & 31.11 &34.54
	     & 35.11 & 32.75 &33.89
	     & 35.61 & 31.09 & 33.20
\\
	     
	     \midrule
	     \multicolumn{13}{l}{\textbf{Entropy-based (Stddev)}} \\
	     \midrule
	     
	     LSTM 
	     & \textbf{67.56} & 40.30 & 50.49
	     & 68.47 & 39.28 & 49.92
	     & 66.52 & 41.10 & 50.81
	     & 51.62 & 42.30 & 46.50\\

	     Transformer 
	     & 
	     66.98 &44.73 & 53.64
	     & 67.72 &40.63& 50.79 
	     & \textbf{66.62} & 42.18 & 51.66
	     & 51.12 & 42.19 & 46.22

 \\
	     \midrule
	     
	     \multicolumn{13}{l}{\textbf{Subword segmental models}} \\
	     \midrule
	     SSLM
	     & 30.93 & \textbf{81.33} & 44.82
	     & 36.17 & \textbf{80.67} & 49.95
	     & 37.28 & \textbf{75.84} &	49.98
	     &  32.54 & \textbf{61.11} & 42.47  \\
	     Word-level 
	     & 65.54 & 50.74 & \textbf{57.20}
	     & \textbf{71.93} & 50.72 & \textbf{59.49}
	     & 61.50 & 53.99 & \textbf{57.50}
	     & \textbf{52.99} & 52.90 & \textbf{52.95} \\
		\bottomrule
	\end{tabular}
	\captionof{table}{Morpheme \emph{boundary} identification (MBI) metrics averaged over the annotated evaluation set.} 	\label{ums2}
\end{figure*}

\section{Unsupervised Morphological Segmentation} \label{sec_ums}

We evaluate our SSLM on unsupervised morphological segmentation (UMS), a challenging task for morphologically rich languages \citep{poon-etal-2009-unsupervised, eskander-etal-2019-unsupervised, unsupervisedembeddings}. This tests to what extent our model discovers morphemes as linguistic units.

\subsection{Data}

We evaluate our models on morphologically annotated data from the Annotated Text Corpora\footnote{Datasets are available at \url{https://repo.sadilar.org/handle/20.500.12185/7}} released by the National Center for Human Technology (NCHLT) in South Africa \cite{eiselen-puttkammer-2014-developing}. 
The dataset required some preprocessing for our use, which we detail in appendix \ref{appendix_morph}. 
We evaluate our models on the test sets for isiXhosa, isiZulu, isiNdebele, and Siswati. Each test set  consists of around 3500 words of free text and morphological analyses of all the words.

\subsection{Models}

\paragraph{Baselines}
Morfessor \cite{creutz-etal-2007-morfessor} is a widely used family of UMS algorithms. We use the Morfessor Baseline model, which is trained on unsegmented words
and based on minimum description length. 
\citet{smit-etal-2014-morfessor} notes that Morfessor tends to undersegment when trained on large corpora.
To evaluate the true potential of Morfessor we view the dataset size as a hyperparameter. We train Morfessor on several subcorpora of our LM training sets, at different orders of dataset size. We report the results of Morfessor trained on a subset of 10k tokens, which gave the best performance.   

\paragraph{Entropy-based segmenter}
We also implemented a character-level entropy-based segmenter, based on the work of \citet{mzamo-etal-2019-towards}. Their approach consists of training a character-level LM and using the entropy of its probability distribution to predict word segment boundaries. The conditional entropy of $x_i$ in a sequence $\mathbf{x}$ is defined as
\begin{align} \label{entropy}
    H(x_i|\mathbf{x_{<i}}) = - \sum_{x \in V} p(x|\mathbf{x_{<i}}) \log p(x|\mathbf{x_{<i}}),
\end{align}
where $V$ is the character vocabulary. 
The entropy-based segmenter splits words at positions where conditional entropy is high. The motivation behind this is that model uncertainty (entropy) will decrease inside a morpheme and increase at morpheme boundaries, where the next character is harder to predict \citep{elman-1990-finding}. 

\citet{mzamo-etal-2019-towards,mzamo-etal-2019-evaluation} train n-gram and bi-LSTM LMs for isiXhosa, while \citet{moeng-etal-2021-canonical} train left-to-right and right-to-left LSTM LMs for isiXhosa, isiZulu, isiNdebele, and Siswati. Both these works experimented with different entropy-based criteria for segmentation, including segmenting on entropy increases, comparing character entropy to mean word entropy, and using thresholds. We use the character-level LSTM and Transformer LMs trained as baselines in \S\ref{sect4_lm}. These LMs were tuned for validation BPC, not segmentation accuracy, since we are applying them as fully \emph{unsupervised} segmenters. We experimented with 3 entropy-based criteria:

\begin{itemize}
\setlength\itemsep{0em}
    \item \textbf{Spike:} Predict subword boundary where entropy increases and then decreases.

\item\textbf{Increase:} Predict subword boundary where entropy increases.

\item\textbf{Stddev:} Predict subword boundary where entropy exceeds one standard deviation greater than the mean sequence entropy.
\end{itemize}

\paragraph{SSLM}

\begin{figure*}%
    \centering
    \subfloat{{\includegraphics[width=7cm]{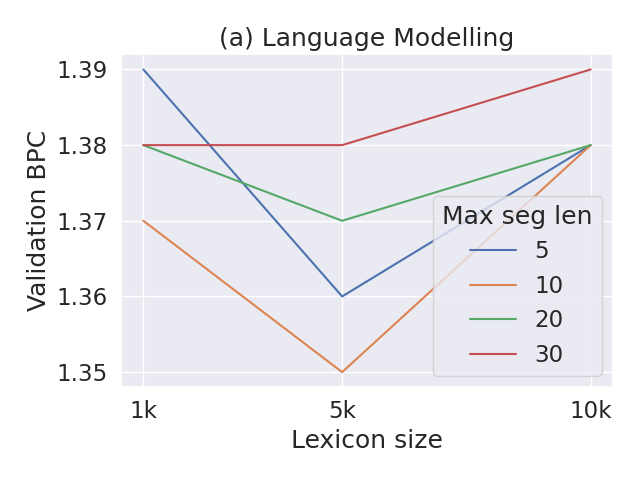} }}%
    \,\,\,
    \subfloat{{\includegraphics[width=7cm]{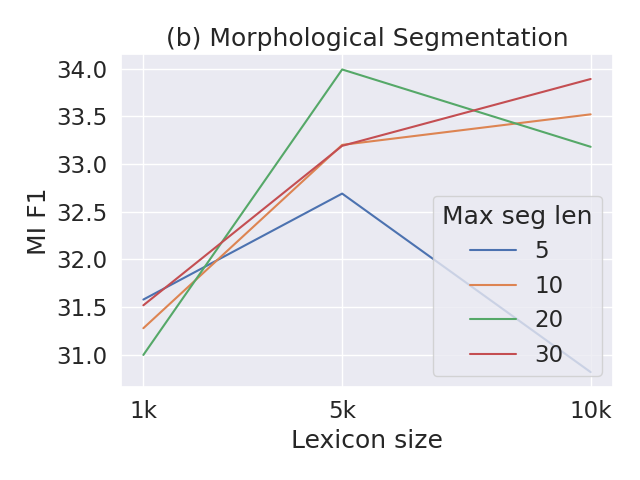} }}%
    
    \vspace{-0.2cm}
    \caption{Comparing isiNdebele SSLM performance across varied lexicon sizes and maximum segment lengths.}%
    \label{lexicon_analysis}%
\end{figure*}

To apply our SSLMs as segmenters we compute the segmentations that maximise the likelihood of a sentence, using the Viterbi algorithm.
For each language we evaluate 2 subword segmental models. 
First we consider the models in \S\ref{sect4_lm}, trained as long-range SSLMs and carrying hidden states between batches.

Second, we introduce a new set of models trained on single words in isolation. These models (which we refer to as our word-level models) are not LMs. 
They are trained on the same datasets as our long-range SSLMs, but they process one word at a time without any surrounding context. This removes the need to model long-range linguistic dependencies, allowing the subword segmental model to focus on the short-range task of word prediction and segmentation. 
We added these models because it is a common approach for UMS algorithms (e.g. Morfessor) to operate on the word-level. We wanted to investigate the morphological segmentation abilities of our subword segmental approach when uncoupled from the task of long-range language modelling.
We tune the word-level models on validation BPC and not on segmentation accuracy, so all our models are fully unsupervised with respect to morphological segmentation.

\subsection{Results}

We report precision, recall, and F1 scores on morpheme identification (MI) and morpheme boundary identification (MBI). 
MBI is the standard measure of morphological segmentation accuracy, but we include MI because it reflects to what extent our models ``discover'' morphemes.
In MI a morpheme is correctly identified if it is among the subwords a word is segmented into. Table \ref{ums1} shows a consistent pattern in the results. The subword segmental models outperform the baselines on all the languages. Our word-level models obtain the best MI precision and F1 scores, while the SSLMs generally obtain better recall. In MBI the goal is to correctly classify whether two consecutive characters are separated by a morpheme boundary. The results in Table \ref{ums2} reveal the same pattern as for MI, although for some languages the entropy-based segmenters obtain greater precision than any of the SSLMs. 

Our word-level subword segmetal models do particularly well, emerging as strong unsupervised morphological segmenters. The SSLMs might not be state-of-the-art segmenters, but still outperform segmenters like BPE and ULM on F1 scores. Surprisingly, Morfessor is the weakest among all the models. On the other hand, the entropy-based segmenters achieve consistently impressive results, confirming it as an effective approach to UMS. We only include results for the \textbf{Stddev} model here, since it is the strongest entropy-based segmenter overall.
The results for all the entropy-based segmenters are included in the appendix.

MBI reveals the type of errors made by our models. The SSLM obtains high recall and low precision, indicating that the model is identifying a large proportion of morpheme boundaries, but often predicting boundaries where there aren't.
This reflects a tendency to over-segment, which might be explained by the low-resource setting. For an LM to utilise long subwords it would have to be exposed to sufficient examples of its use. This might not be possible with smaller training sets, so the model relies on shorter segments instead. 

Nevertheless, the SSLM outperforms BPE and ULM on UMS. The greater linguistic plausibility of its segmentations might contribute to its strength as a LM in \S\ref{sect4_lm}. Letting the model learn which segmentations for language modelling results in the model discovering, to some extent, morphemes as linguistic units.

%% file: 6analysis.tex
\begin{figure*}[t] \small
    \centering
	
	\begin{tabular}{p{1.7cm}|p{12cm}} 
	    \multicolumn{2}{c}{}\\	   
		\toprule
		 \textbf{Sentence} & Sibuye sithokoze khulu kwamanikelela emphakathini weentjhabatjhaba ngesekelo labo elinganakuzaza emzabalazweni wethu.\\
		 \midrule
		  \textbf{Morphemes} & Si-buy-e si-thokoz-e khulu k-w-amanik-elel-a e-m-phakath-ini weentjhabatjhaba nge-sekelo labo eli-nga-nakuzaza  e-mzabalazw-eni  w-ethu. \\
	    
	     \midrule
	     
	     \textbf{SSLM} & \colorcorrect{Si}-\colorcorrect{buy}-\colorcorrect{e} s-i-t-h-oko-z-e k-hulu kwam-a-nikele-l-\colorcorrect{a} \colorcorrect{e}-\colorcorrect{m}-\colorcorrect{phakath}-i-n-i ween-tjhaba-tjhab-a n-g-e-\colorcorrect{sekelo} l-a-b-o e-l-i-ngana-kuz-az-a \colorcorrect{e}-mz-abal-az-w-e-n-i \colorcorrect{w}-e-thu. \\
	     \midrule
	     \textbf{BPE} & \colorcorrect{Si}-bu-ye \colorcorrect{si}-tho-ko-ze \colorcorrect{khulu} kwa-m-ani-k-elela em-phaka-thini ween-tjhaba-tjhaba \colorcorrect{nge}-se-k-elo la-bo \colorcorrect{eli}-ng-ana-ku-za-za em-za-bala-z-weni we-thu. \\
	     \midrule
	     \textbf{ULM} & \colorcorrect{Si}-bu-ye \colorcorrect{si}-tho-ko-ze \colorcorrect{khulu} kwama-nikele-la emphakathin-i w-eentjhabatjhab-a \colorcorrect{nge}-se-ke-lo la-bo e-lingana-ku-za-za em-za-ba-la-zwe-ni  we-thu. \\
		\bottomrule
	\end{tabular}
	\captionof{table}{The output of subword segmenters compared to the annotated morphological segmentation of an isiNbedele sentence. Correctly identified morphemes are indicated in green.} 	\label{sentence_examples}
\end{figure*}

\section{Analysis} \label{sec6_analysis}


We analyse the effect of hyperparameter and architectural choices on the performance of our SSLMs. This section does not report findings for our word-level models, since we are primarily interested in  investigating which model components contribute to the success of our long-range SSLMs (in terms of language modelling and morphological segmentation).
\citet{kawakami-etal-2019-learning} found two components to be crucial to the success of their segmental LM for Chinese word discovery: the lexicon and expected length regularisation. The former stores frequent subwords and the latter introduces a regularisation term to the training objective that encourages shorter segments.

Contrary to \citet{kawakami-etal-2019-learning}, we did not find length regularisation to be useful.
This is because our datasets are much larger than those used in their ablation studies (they use the Brent corpus of 27k words). When a segmental model is trained on a small dataset, it overfits by copying long segments from the lexicon.
Length regularisation prevents overfitting by biasing the model towards shorter segments. This is not a problem on larger datasets like ours, because the lexicon cannot cover all possible long segments in the corpus. 
In fact, our model tends to over-segment rather than under-segment. Table \ref{avg_segment} confirms this, showing that SSLM subwords are on average much shorter than morphemes.
It is also evident in Table \ref{sentence_examples}, where we show a segmented isiNdebele sentence.
The SSLM often over-segments, but sometimes its segmentations are more accurate because of its tendency towards shorter segments.
In the examples BPE and ULM fail to identify any of the 1-character morphemes, while the SSLM identifies several.


\begin{figure}[t] 
    \centering
	\begin{tabular}{lcccc} 
		\toprule
	     \textbf{Language} & \textbf{xh}& \textbf{zu}& \textbf{nr}& \textbf{ss} \\
	     \midrule
	     Morphemes& 2.93 & 2.86& 3.03 & 3.45\\
	     SSLM segments & 1.43 & 1.48 & 1.64& 2.24\\
		\bottomrule
	\end{tabular}
	\captionof{table}{Average subword length on UMS test sets.} 	\label{avg_segment}
\end{figure}

The lexicon proved to be essential for the SSLM. During tuning it consistently improved validation BPC.
The average lexicon coefficient ($1-g_t$ in equation \ref{mixture}) of the isiNdebele SSLM on the LM test set was 0.27, indicating a reliance on the lexicon for subword generation.
We analyse two lexicon-related hyperparameters: lexicon size and maximum segment length.
Figure \ref{lexicon_analysis} compares the performance of isiNdebele SSLMs across lexicon sizes and maximum segment lengths.
Figure \ref{lexicon_analysis} (a) plots intrinsic LM performance. Smaller lexicon sizes improve LM performance up to a point, with 5k subwords being optimal. A maximum segment length of 10 characters achieves optimal performance across all lexicon sizes, striking a balance between memorising long segments where possible, and otherwise relying on short subwords.

Figure \ref{lexicon_analysis} (b) plots UMS performance, where the picture is less clear-cut.
Since our model is an \emph{unsupervised} morphological segmenter, we only considered LM performance (validation BPC) when tuning and selecting our final models. Figure \ref{lexicon_analysis} shows that optimal LM performance does not necessarily imply optimal UMS performance. Longer maximum segment lengths tend to improve UMS performance. Biasing the model towards longer segments reduces the over-segmentation problem, but relinquishes some LM performance. 
However, there is at least some correlation between LM and UMS performance, and the model selected on LM performance is not far off optimal UMS accuracy.

%% file: 7related_work.tex
\section{Related Work}

A few African languages have been included in large multilingual LMs, such as mBERT \citep{devlin-etal-2019-bert} and XLM-R \citep{conneau-lample-2019-cross-lingual}.
\citet{ogueji-etal-2021-small} trained AfriBERTa, a multilingual LM trained on 11 African languages.
There has been less work on monolingual LMs for African languages. 
\citet{ralethe-2020-adaptation} trained AfriBERT, a masked LM for Afrikaans.
\citet{mesham-etal-2021-low} trained BPE-based autoregressive LMs for isiZulu and Sepedi. 
\citet{nzeyimana-rubungo-2022-kinyabert} proposed KinyaBERT, a masked LM for Kinyarwanda with a two-tier neural architecture that incorporates a morphological analyzer. 



%% file: 8conclusion.tex
\section{Conclusion}

In this paper we proposed subword segmental language modelling (SSLM), an approach that unifies language modelling and subword segmentation.
We showed that SSLM improves intrinsic LM performance for low-resource agglutinative languages, while yielding subwords that approximate morphemes better than previous approaches. 
As opposed to most neural model architectures in NLP research, which are either language-agnostic or overfit to high-resource languages, our model is designed to suit agglutinative languages like the Nguni languages of South Africa.
Our results show that learning subword segmentation in training overcomes some of the limitations of existing subword segmenters.
For future work, the SSLM could be applied to downstream NLP tasks suited to its autoregressive architecture, such as text generation. More generally, the idea of learning subword segmentation during training could be adapted to other NLP models and tasks.

%% file: 9limitations.tex


\section*{Limitations}

We evaluate our model on languages from a single language group - the Nguni languages. Our findings might not hold for languages with different types of morphological complexity (e.g. fusional languages, where segmentation is difficult because morphemes are fused together). The SSLM achieved consistently good LM performance across all four Nguni languages, but we had to tune the lexicon size and maximum segment length separately for each language. These optimal hyperparameter values varied across languages and would have to be tuned from scratch for new languages.

Our subword segmental approach is able to improve over all baselines as a morphological segmenter, but only if we train it as a word-level sequence model. The SSLM outperforms standard segmenters like BPE and ULM, but performs worse than our entropy-based baselines on F1 scores. This shows that there is a deterioration in segmentation performance because the SSLM is required to model long-range linguistic dependencies - the model tends to over-segment words. 
We only evaluate our segmentations with automatic evaluation metrics, which provides a rigid, morpheme-based perspective on the segmentation quality. It would be ideal to also include human evaluations of the linguistic plausibility of segmentations. 

\section*{Ethical Considerations}
We release LM datasets for 4 Nguni languages, consisting of free text split into train/validation/test sets. Our datasets are compilations of existing, publicly available datasets. The datasets we sourced are listed in Table \ref{datasets_lang} in the appendix. As outlined in section \ref{datasubsction}, we took certain steps to ensure that the datasets we sourced were of reasonable quality. 

Nevertheless, since the source datasets were ori-ginally scraped from the web, we acknowledge that we do not avoid all the pitfalls of large scale data collection for low-resource languages (most notably, the presence of text in other languages). Furthermore, most of the data is sourced from South African government publications, so they are domain-specific to some extent. The texts cover diverse topics, but generally fall within the categories and style expected of government publications.

\section*{Acknowledgements}
This work is based on research supported in part by the National Research Foundation of South Africa (Grant Number: 129850) and the South African Centre for High Performance Computing.  
Computations were performed using facilities provided by the University of Cape Town’s ICTS High Performance Computing team: \url{hpc.uct.ac.za}.
Francois Meyer is supported by the Hasso Plattner Institute for Digital Engineering, through the HPI Research School at the University of Cape Town. 


%% file: appendix.tex
\appendix



\section{SSLM Training}
\label{appendix_hyperparameters}

We tuned our SSLMs to optimise BPC on our LM validation sets. We used the Adam optimizer \cite{kingma-ba-2015-adam}, halving the learning rate if validation loss failed to improve for 3 epochs and stopping when no validation loss improvement occurred for 6 epochs. We used several standard regularisation techniques, including dropout (on all except recurrent layers), weight decay, and gradient clipping.
Table \ref{hyperparams_common} shows the hyperparameter settings we used for our subword segmental models and baseline LMs. The optimal lexicon size and maximum segment length varied across the languages, as shown in Table \ref{hyperparams_lang}.

We trained our models on virtually partitioned instances of NVIDIA A100 GPUs with 3 compute units and 20GB memory. The isiXhosa and isiZulu long-range SSLMs converged after about 40 epochs of the training corpus, taking 3 days to train. The isiNdebele and Siswati long-range SSLMs converged after 30 epochs, taking 10 hours to train.
The word-level SSLMs trained quite slowly, since each word is processed as an individual sequence. Therefore we trained our isiXhosa and isiZulu models on 500k-word subcorpora of the LM training sets (matching the sizes of the isiNdebele and Siswati datasets). These converged before 20 epochs, taking 10 hours to train. Segmenting the morphological evaluation data took less than a few minutes on a laptop computer, since the Viterbi algorithm is computationally efficient.

\section{UMS Data}
\label{appendix_morph} 

The data consists of a train/test set of free text in which words have been morphologically analysed.
Words are segmented into their \emph{canonical segmentations} i.e. standardised morphemes that do not necessarily correspond to word substrings \cite{cotterell-etal-2016-joint}.
For example, the canonical segmentation of the isiXhosa word ``yedwa'' is ``ya-i-dwa''. 
Since our SSLM segments words into substrings, we require \emph{surface segmentations} (segments correspond to substrings) for evaluation.
Most of the segmentations can be used as is, because the canonical and surface segmentations are identical. Where the segmentations differ, we use the scripts made available by \citet{moeng-etal-2021-canonical} to map canonical segmentations to surface segmentations. They employ a heuristic approach based on the Levenshtein distance minimal edit operations to map from the de-segmented canonical form to the surface form. They also filter out tokens that are unsuitable for morphological segmentation.


\begin{figure}[h]
    \centering
	\begin{tabular}{lcc} 
		\toprule
	     \textbf{Model} & \textbf{Lexicon size}& \textbf{Max seg len} \\
	     \midrule
	     \multicolumn{3}{l}{\textbf{Long-range SSLM}} \\
	     \midrule
	     isiXhosa & 10k & 5    \\
	     isiZulu & 10k & 5\\
	     isiNdebele & 5k & 10\\
	     Siswati & 10k & 20\\
	     \midrule
	     \multicolumn{3}{l}{\textbf{Word-level SSLM}} \\
	     \midrule
	     isiXhosa & 10k & 10    \\
	     isiZulu & 5k & 20\\
	     isiNdebele & 10k & 10\\
	     Siswati & 5k & 20\\
		\bottomrule
	\end{tabular}
	\captionof{table}{Lexicon hyperparameters for our SSLMs.} 	\label{hyperparams_lang}
\end{figure}
	     
	     
	     

\begin{figure*}[t]
    \centering
	\begin{tabular}{lcccc} 
		\toprule
		     & \multicolumn{2}{c}{\textbf{Subword segmental models}} & \multicolumn{2}{c}{\textbf{Baseline models}} \\
	     \midrule
	     \textbf{Hyperparameter} & \textbf{Long-range}& \textbf{Word-level} & \textbf{LSTM}&
	     \textbf{Transformer}\\
	     \midrule
	     Attention heads &  &  & & 4/8*  \\
	     LSTM layers & 3 & 1  & 3 & 3   \\
	     Embedding size & 512 & 512 & 128/512* & 512\\
	     Hidden size & 1024 & 1024 & 1024 & 1024\\
	     
	     Learning rate & 0.001 & 0.005 & 0.001 & 0.001\\
	     Dropout & 0.5 & 0.2  & 0.2 & 0.1\\
	     
	     Batch size & 64 & 16 & 64 & 64\\
	     Sequence & 120 chars & 1 word  & 120 chars & 120 chars \\
	     
	     Weight decay & 1e-5 & 1e-5 & 1e-5 & 1e-5  \\
	     Gradient clip & 1.0 & 1.0 & 1.0 & 1.0  \\
		\bottomrule
	\end{tabular}
	\captionof{table}{Hyperparameter settings for all our models, with * indicating where the optimal hyperparameter value (based on validation BPC) depended on the language. For embedding size, 128 was optimal for isiXhosa and Siswati, while 512 was optimal for isiZulu and isiNdebele. The Transformer models had 8 attention heads for isiXhosa and isiZulu, and 4 for isiNdebele and Siswati.} 	\label{hyperparams_common}
\end{figure*}

\begin{figure*}[t]
    \centering
	\begin{tabular}{lccc} 
		\toprule
	     \textbf{Data set} & \textbf{Type}& \textbf{Source}\\
	     \midrule
	     \multicolumn{3}{l}{\textbf{isiXhosa}} \\
	     \midrule
	     NCHLT Text & monolingual & South African government websites \\
	     SADiLaR Monolingual & monolingual & South African government websites\\
	     Navy Corpus & parallel & South African government websites\\
	     \midrule
	     \multicolumn{3}{l}{\textbf{isiZulu}} \\
	     \midrule
	     NCHLT Text & monolingual &South African government websites \\
	     Autshumato & parallel & South African government websites\\
	     Isolezwe News Corpus & monolingual & news articles\\
	     \midrule
	     \multicolumn{3}{l}{\textbf{isiNdebele}} \\
	     \midrule
	     NCHLT Text & monolingual & South African government websites \\
	     \midrule
	     \multicolumn{3}{l}{\textbf{Siswati}} \\
	     \midrule
	     NCHLT Text & monolingual & South African government websites \\
		\bottomrule
	\end{tabular}
	\captionof{table}{Our language modelling data sets were compiled from these publicly available data sets. We split the individual corpora into train/validation/test sets before combining them respectively into one train/validation/test data set. This ensured that the individual corpora are equally distributed in the training and evaluation sets.} 	\label{datasets_lang}
\end{figure*}


 
\begin{figure*}[t]\small
    \centering
	\begin{tabular}{l |ccc| ccc |ccc| ccc} 
		\toprule
	     & \multicolumn{3}{c}{\textbf{xh}} & \multicolumn{3}{c}{\textbf{zu}} & \multicolumn{3}{c}{\textbf{nr}} & \multicolumn{3}{c}{\textbf{ss}} \\
	     \midrule 
	     & P & R & F1 & P & R & F1 & P & R & F1 & P & R & F1\\

	     \midrule
	     \multicolumn{13}{l}{\textbf{Char-LSTM Entropy}} \\
	     \midrule
	     Spike & 
	     34.71 & 36.97 & 35.81& 33.56 & 34.36 & 33.95
 & 29.12& 32.62 & 30.77 &
28.94&35.38&31.83\\
	     Increase & 
	     27.64 & 38.14 & 32.06
 & 29.21 & 37.52 & 32.85
 & 25.90 & \textbf{36.36} & 30.25
 & 24.32&\textbf{37.01}&29.35\\
	     Stddev & 
	     40.99 & 30.43 & 34.93
 & \textbf{39.26} & 28.23 & 32.85
 & 38.61 & 29.01 & 33.13
 & \textbf{33.95}&30.40&\textbf{32.07}
\\
	     \midrule
	     \multicolumn{13}{l}{\textbf{Char-Transformer Entropy}} \\
	     \midrule
	     Spike & 
	     34.80 & 37.84 & 36.26
        & 33.22 & 34.83 & \textbf{34.00}
        & 27.03 & 31.10 & 28.92
 & 25.82&33.12&29.02\\
	     Increase & 
	     29.65 & \textbf{40.00} & 34.06
 & 29.83 & \textbf{38.47} & 33.60
 &  24.89 & 35.03 & 29.10
 & 22.86& 35.91&27.93\\
	     Stddev & 
	     \textbf{42.99} & 33.92 & \textbf{37.92}&  39.24 & 28.93 &33.31&
	     \textbf{38.93} &29.65 &\textbf{33.66} &
	     33.16&29.74&31.35\\

		\bottomrule
	\end{tabular}
	\captionof{table}{Morpheme identification performance metrics for all entropy-based models.} 
\end{figure*}

\begin{figure*}[t]\small
    \centering
	\begin{tabular}{l |ccc| ccc |ccc| ccc} 
		\toprule
	     & \multicolumn{3}{c}{\textbf{xh}} & \multicolumn{3}{c}{\textbf{zu}} & \multicolumn{3}{c}{\textbf{nr}} & \multicolumn{3}{c}{\textbf{ss}} \\
	     \midrule 
	     & P & R & F1 & P & R & F1 & P & R & F1 & P & R & F1\\
	     \midrule

	     \multicolumn{13}{l}{\textbf{Char-LSTM Entropy}} \\
	     \midrule
	     Spike 
	     & 51.08 &56.25 & 53.54	 & 53.08 & 54.98 & \textbf{54.01}
	     & 46.36 &54.77 & 50.22
	     & 41.57 & 56.61 & \textbf{47.93}

\\
	     Increase 
	     & 40.15 & 63.61 & 49.23
	     & 44.62 & 63.41 & 52.38
	     & 39.31 & \textbf{63.01} &48.42
	     & 35.48 & \textbf{65.27} &45.97

\\
	     Stddev 
	     & \textbf{67.56} & 40.30 & 50.49
	     & \textbf{68.47} & 39.28 & 49.92
	     & 66.52 & 41.10 & 50.81
	     & \textbf{51.62} & 42.30 & 46.50

\\
	     \midrule
	     \multicolumn{13}{l}{\textbf{Char-Transformer Entropy}} \\
	     \midrule
	     Spike 
	     & 50.40 & 57.24 & 53.60
	     & 52.11 & 55.91 & 53.94
	     & 44.35 & 54.40 &48.86
	     & 38.03 & 55.52 & 45.14

\\
	     Increase 
	     & 42.12 & \textbf{64.68} & 51.02
	     & 45.19 &\textbf{64.59}& 53.17
	     & 38.36 & 61.62 & 47.29
	     & 33.16 & 63.73 & 43.62

\\
	     Stddev 
	     & 
	     66.98 &44.73 &\textbf{53.64}
	     & 67.72 &40.63& 50.79 
	     & \textbf{66.62} & 42.18 & \textbf{51.66}
	     & 51.12 & 42.19 & 46.22\\

		\bottomrule
	\end{tabular}
	\captionof{table}{Morpheme boundary identification performance metrics for all entropy-based models.} 	
\end{figure*}

\begin{figure*}[t] \small
    \centering
	\begin{tabular}{p{1.7cm}|p{12cm}} 
	    \multicolumn{2}{c}{\textbf{(a) isiXhosa sentence segmentation}} \\
		\toprule
		 \textbf{Sentence} & Siphinda kwakhona umbulelo wethu osuka emazantsi entliziyo kwabezizwe ngezizwe ngenkxaso yabo engagungqiyo ekuxhaseni umzabalazo wethu.\\
		 \midrule
		  \textbf{Morphemes} & Si-phind-a kwa-khona u-m-bulelo w-ethu  o-suk-a emazantsi e-n-tliziyo kwa-bezizwe  nge-zi-zwe  ngenkxaso y-abo engagungqiyo  e-ku-xhas-eni u-m-zabalazo  w-ethu.  \\
	    
	     \midrule
	     
	     \textbf{SSLM} &  S-i-phin-d-\colorcorrect{a} kwak-hon-a \colorcorrect{u}-\colorcorrect{m}-bule-l-o \colorcorrect{w}-e-thu osuk-\colorcorrect{a} e-m-a-zant-s-i \colorcorrect{e}-\colorcorrect{n}-tl-iziy-o k-w-a-b-e-z-i-zw-e ngezi-zw-e n-g-e-nkxa-s-o \colorcorrect{y}-a-b-o e-ngag-ungqi-yo \colorcorrect{e}-k-u-\colorcorrect{xhas}-e-n-i \colorcorrect{u}-\colorcorrect{m}-zab-alaz-o \colorcorrect{w}-e-thu.\\
\midrule
	     \textbf{BPE} & \colorcorrect{Si}-phin-da \colorcorrect{kwa}-\colorcorrect{khona} um-bu-lelo we-thu \colorcorrect{o}-su-ka ema-zantsi ent-li-zi-yo \colorcorrect{kwa}-b-ezi-zwe ngezi-\colorcorrect{zwe} ngen-kxaso ya-bo enga-gu-ng-qi-yo eku-xha-s-\colorcorrect{eni} um-za-bala-zo we-thu. \\
	    \midrule
	     \textbf{ULM} & \colorcorrect{Si}-phi-nda \colorcorrect{kwa}-\colorcorrect{khona} \colorcorrect{u}-mbu-lelo we-thu \colorcorrect{o}-suka e-ma-za-nts-i \colorcorrect{e}-nt-li-zi-yo \colorcorrect{kwa}-be-zi-z-we \colorcorrect{nge}-\colorcorrect{zi}-z-we nge-nkxaso ya-bo e-nga-gu-ngq-i-yo \colorcorrect{e}-\colorcorrect{ku}-xh-a-se-ni um-za-ba-la-zo we-thu\\
		\bottomrule
	\end{tabular}
	
	\vspace{0.5cm}
	
	\begin{tabular}{p{1.7cm}|p{12cm}} 
	    \multicolumn{2}{c}{\textbf{(b) isiZulu sentence segmentation}} \\
		\toprule
		 \textbf{Sentence} & Siyaphinda sibonga siyanconcoza emphakathini womhlaba ngokuseseka kwawo emzabalazweni wethu.\\
		 \midrule
		  \textbf{Morphemes} & Si-ya-phind-a si-bong-a si-ya-nconcoz-a e-m-phakath-ini wo-m-hlaba n-gokusesek-a kwa-wo e-mzabalazw-eni w-ethu. \\
	    
	     \midrule
	     
	     \textbf{SSLM} &  S-i-yaph-inda s-i-	    \colorcorrect{bong}-\colorcorrect{a} siya-nco-nco-z-\colorcorrect{a} \colorcorrect{e}-mphak-a-t-h-i-n-i w-o-m-\colorcorrect{hlaba} \colorcorrect{n}-g-o-k-u-s-eseka k-w-a-w-o emz-abala-z-w-e-n-i wethu.\\
\midrule
	     \textbf{BPE} & \colorcorrect{Si}-\colorcorrect{ya}-phi-nda \colorcorrect{si}-bo-nga \colorcorrect{si}-\colorcorrect{ya}-n-co-n-co-za em-phakathi-ni \colorcorrect{wo}-m-\colorcorrect{hlaba} ngoku-se-se-ka \colorcorrect{kwa}-\colorcorrect{wo} em-za-bala-zweni we-thu. \\
	    \midrule
	     \textbf{ULM} & \colorcorrect{Si}-\colorcorrect{ya}-phi-nda \colorcorrect{si}-bo-nga \colorcorrect{si}-\colorcorrect{ya}-n-co-n-co-za emphakathini \colorcorrect{wo}-mhlaba ngoku-se-se-ka \colorcorrect{kwa}-\colorcorrect{wo} em-za-ba-la-zwe-ni we-thu. \\
		\bottomrule
	\end{tabular}

	\vspace{0.5cm}
	
	\begin{tabular}{p{1.7cm}|p{12cm}} 
	    \multicolumn{2}{c}{\textbf{(c) Siswati sentence segmentation}}\\
		\toprule
		 \textbf{Sentence} &  Sendlulisa kubonga kwetfu kummango wemave emhlaba ngekwesekela umzabalazo wetfu ngendlela lengenakunyakatiswa.\\
		 \midrule
		  \textbf{Morphemes} & S-endlulis-a ku-bong-a kwetfu  ku-mmango wemave emhlaba  ngekwesekela u-mzabalazo  wetfu  nge-n-dlela  le-n-genakunyakatiswa.\\
	    
	     \midrule
	     
	     \textbf{SSLM} & \colorcorrect{S}-e-n-dlulis-\colorcorrect{a} k-u-\colorcorrect{bong}-\colorcorrect{a} kw-e-tfu k-u-m-mango w-e-mave \colorcorrect{emhlaba} n-g-e-k-w-e-sekel-a \colorcorrect{u}-m-zaba-laz-o w-e-tfu ngendle-l-a l-e-ngena-ku-n-yaka-t-isw-a\\
	     \midrule
	     \textbf{BPE} &  \colorcorrect{S}-en-dlu-lisa kubo-nga  \colorcorrect{kwetfu} kum-ma-ngo wema-ve em-hlaba ngekwe-se-kela um-za-bala-zo we-tfu ngendlela  \colorcorrect{le}-ngen-aku-nya-kati-swa. \\
	     \midrule
	     \textbf{ULM} & Se-ndlu-lisa kubo-nga kwe-tfu \colorcorrect{ku}-m-ma-ngo we-mave e-mhlaba ngekwe-se-kela um-za-ba-la-z-o we-tfu ngendlela \colorcorrect{le}-ngena-ku-nya-kati-swa.  \\
		\bottomrule
	\end{tabular}
	
	\captionof{table}{
	The output of subword segmenters compared to the annotated morphological segmentation of Nguni language sentences. Correctly identified morphemes are indicated in green.} 	
	
\end{figure*}